\documentclass[journal]{IEEEtran}
\usepackage[pdftex]{graphicx}
\pdfoutput=1

\usepackage{multirow}
\usepackage{multicol}
\usepackage{booktabs}
\usepackage{graphicx}
\usepackage{lineno}
\usepackage{array}
\usepackage{color}
\usepackage{makecell}
\usepackage[colorlinks,linkcolor=red,anchorcolor=green,citecolor=green]{hyperref}
\usepackage{cite}

\ifCLASSINFOpdf
\else
\fi
\usepackage{amsmath}
\usepackage{url}

\usepackage{algorithm}
\usepackage{algpseudocode}
\usepackage{amsmath}
\usepackage{graphics}
\usepackage{epsfig}
\usepackage{breqn}

\begin{document}
%
\title{Perceptual Multi-Exposure Fusion}
%
%
%

\author{Xiaoning~Liu
\thanks{X. Liu is with the School of Information and Communication Engineering, University of Electronic Science and Technology
	of China, Chengdu, 611731, China. E-mail: liuxiaoning2016@sina.com}
}
\maketitle

\begin{abstract}
As an ever-increasing demand for high dynamic range (HDR) scene shooting, multi-exposure image fusion (MEF) technology has abounded. In recent years, multi-scale exposure fusion approaches based on detail-enhancement have led the way for improvement in highlight and shadow details. Most of such methods, however, are too computationally expensive to be deployed on mobile devices. This paper presents a perceptual multi-exposure fusion method that not just ensures fine shadow/highlight details but with lower complexity than detail-enhanced methods. We analyze the potential defects of three classical exposure measures in lieu of using detail-enhancement component and improve two of them, namely adaptive Well-exposedness (AWE) and the gradient of color images (3-D gradient). AWE designed in YCbCr color space considers the difference between varying exposure images. 3-D gradient is employed to extract fine details. We build a large-scale multi-exposure benchmark dataset suitable for static scenes, which contains 167 image sequences all told. Experiments on the constructed dataset demonstrate that the proposed method exceeds existing eight state-of-the-art approaches in terms of visually and MEF-SSIM value. Moreover, our approach can achieve a better improvement for current image enhancement techniques, ensuring fine detail in bright light.
\end{abstract}


%
\IEEEpeerreviewmaketitle

\section{Introduction}
%
%
%
%
\IEEEPARstart{H}{igh} dynamic range (HDR) imaging technique could make the image captured in extremely bright or dark condition crispy and faithfully accessible to a real-world scene. It has increasingly catered to mobile devices, videos, autonomous vehicles and so on \cite{Mohit_2013}, \cite{Ma_2015}. Unfortunately, its wide range of applications is affected by expensive equipment cost and visualizing on a standard display with a limited dynamic range which usually relies on tone mapping operations \cite{Gu_2016, Durand_2002, Li_2018, Li_2014, Hadizadeh_2017, Wang_2015}. To mitigate these limitations, multi-exposure image fusion (MEF) technology, also termed exposure bracketing \cite{Im_2018}, merges multiple exposure images captured from same scenes with different exposure time into a spectacular HDR image abounded with desirable detail information. Since MEF technology simplifies the HDR imaging pipeline, it has recently accommodated smart cameras, particularly smartphones. MEF techniques, however, inevitably lead to unwelcome artifacts like ghosting and tearing when encountering moving objects or camera shake \cite{Im_2018, Marcelo_2012, Khademi_2017, Sen_2012}. To overcome this challenge, many HDR deghosting algorithms \cite{Jun_2013, Li_2014_Selectively, Oh_2014, tursun2015state, tursun2016objective, karadjuzovic2017assessment,  gallo2009artifact, Hasinoff_2016} have been proposed. Among them, Tursun $\emph{et al}.$ carried out an in-depth survey of HDR deghosting \cite{tursun2015state} and proposed an objective deghosting quality metric to avoid the bias of subjective evaluations \cite{tursun2016objective}. Although ghost-free methods have made significant headway over the past decade, removing ghosting artifacts is still the greatest challenge to MEF and HDR imaging for dynamic scenes \cite{Kinoshita_2019}. This work assumes input images to be all well-aligned for static scenes. The past two decades have seen a significant amount of work in MEF community. They are generally classified into four categories: multi-scale transform-based methods, statistical model-based methods, patch-based methods, and deep learning-based methods.
\subsection{Multi-Scale Transform-Based Methods}
Initiatively guided by three intrinsic image quality metrics namely contrast, saturation and well-exposedness, Mertens $\emph{et al}.$ \cite{Mertens_2009} constructed the weight maps to blend multiple exposure images in the framework of Laplacian pyramid (LP) \cite{Burt_1983}. Since multi-scale technique can reduce unpleasant halo artifacts around edges and alleviate the seam problem across object boundaries to some extent, multi-scale transform-based MEF approaches \cite{Mertens_2009, Nejati_2017, Kou_2018, Ancuti_2016, Yang_2018}, especially those based on LP \cite{Mertens_2009, Kou_2018, Ancuti_2016, Yang_2018}, have gained ground in popularity. Two novel exposure measures, visibility and consistency in \cite{Zhang_2011}, are developed based on the scrupulous observation that the gradient magnitude will gradually decrease among over/under-exposure areas and the gradient direction will change as the object moves. In order to reduce the loss of detail in multi-scale fusion, Li $\emph{et al}.$ \cite{Li_2012} introduced a new quadratic optimization scheme in the gradient field. The sharper image is finally synthesized by combining the extracted detail with an intermediate image generated by the MEF method \cite{Mertens_2009}. It is unfortunate that the computational efficiency in \cite{Li_2012} cannot deploy mobile devices due to the need of solving a quadratic optimization problem by means of an iterative method. Shen $\emph{et al}.$ \cite{Shen_2014} derived a novel boosting Laplacian pyramid which boosts the structure of detail and base layers, respectively, and designed a hybrid exposure weight. As the optimal weights \cite{Shen_2014} computed by a global optimization may over-smoothing the final weight map, Li $\emph{et al}.$ \cite{Li_2012fast, Li_2013image} used edge-preserving filters, namely recursive filter \cite{Gastal_2011} and guided filter \cite{He_2010}, to refine the resulting weight map under a two-scale framework. 

Furthermore, to ensure high levels of detail with well-controlled artifacts even in low/high-light scenarios, the works in \cite{Li_2017, kou2017multi, kou2017intelligent} recently integrated multi-scale technique with detail-enhanced smoothing pyramid relied on weighted guided image filter \cite{li2014weighted}, gradient domain guided image filter \cite{kou2015gradient}, and fast weighted least square filter \cite{min2014fast}, respectively. Noted that the detail-enhancement technology \cite{Li_2017} is also beneficial to low-light and back-light imaging. Because a fast weighted least square based optimization problem is subjected to a gradient constraint, the speed of detail extraction component in \cite{kou2017intelligent} is significantly faster than that \cite{Li_2012}. Experimental results in \cite{cai2018learning} demonstrated that \cite{kou2017multi} ranks first according to quality metric MEF-SSIM \cite{ma2015perceptual}. Even though \cite{Li_2017, kou2017multi, kou2017intelligent} are capable of high levels of detail in bright and dark light, the complexity of these algorithms may be a hindrance to mobile devices. Lately, in the presence of the multi-scale fusion scheme \cite{Mertens_2009}, a simplified detail enhancement component \cite{wang2019detail} was presented in the YUV color space. 
\subsection{Statistical Model-Based and Patch-Based Methods}
Statistical approaches \cite{shen2011generalized, song2011probabilistic, shen2012qoe} based on perceived quality measures and patch-based methods \cite{ma2015multi, ma2017robust, qin2014robust} have been improved a lot over the past decade. Integrating visual measures, namely local contrast and color information, Shen $\emph{et al}.$ proposed two different frameworks, generalized random walks \cite{shen2011generalized} and hierarchical multivariate Gaussian conditional random field \cite{shen2012qoe}, respectively. Song $\emph{et al}.$ \cite{song2011probabilistic} converted MEF into a maximum posteriori (MAP) framework. Experimental results in \cite{ma2017robust} showed that the final fused image \cite{song2011probabilistic} tends to be noisy due to lack of explicit refinement of weights. To reduce noise and remove ghosting from a dynamic scene, Ma $\emph{et al}.$ \cite{ma2015multi, ma2017robust} originally decomposed each image patch into three conceptually independent components, namely signal strength, signal structure, and mean intensity, and then extracted ones upon patch strength, exposedness and structural consistency measures. The benefits of this novel decomposition abound, one of which is that the direction information of signal structure can guide us to verify structural consistency for producing ghost-free images.
\subsection{Deep Learning-Based Methods}
Lately, MEF approaches \cite{ram2017deepfuse, li2018multi, liu2019variable, zhang2020ifcnn} based on convolutional neural network (CNN) have come a long way from a past featured by hand-rafted features \cite{Mertens_2009, Burt_1983, Nejati_2017, Kou_2018, Ancuti_2016, Yang_2018, Zhang_2011, Li_2012, Shen_2014, Li_2012fast, Li_2013image, Gastal_2011, He_2010, Li_2017, kou2017multi, kou2017intelligent, li2014weighted, kou2015gradient, min2014fast, cai2018learning, ma2015perceptual, wang2019detail, shen2011generalized, song2011probabilistic, shen2012qoe, ma2015multi, ma2017robust, qin2014robust}. Prabhakar $\emph{et al}.$ \cite{ram2017deepfuse} recently proposed a CNN based architecture that actually consists of two methods. One is to train the CNN model with the results selected from two MEF methods \cite{Mertens_2009, Li_2013image} that are considered the top-ranking ones at the time as the ``ground truth". The other is to learn CNN by utilizing a no-reference quality metric MEF-SSIM \cite{ma2015perceptual} as loss function. Considering the lack of ground-truth for MEF, Li $\emph{et al}. $\cite{li2018multi} extracted features from pre-trained CNN on other tasks to calculate weight maps, which is also suitable for dynamic scenes, while the model \cite{ram2017deepfuse} is only for static scenes. A variable augmented neural network \cite{liu2019variable} for de-colorization was designed in the gradient domain. The network is also employed in blending multiple exposure images by revealing the relationship between de-colorization and MEF. The network \cite{liu2019variable}, however, can only feed three exposure images at the same time, which greatly limits its practical application. Zhang $\emph{et al}.$ \cite{zhang2020ifcnn} most recently proposed a multi-modal image fusion framework based on CNN, which takes advantage of two convolutional layers to extract low-level features from multiple input images and blends these features through alternative strategies. \cite{zhang2020ifcnn} offers reasonable texture and detail preservation in the shadow regions, aside from some visually noticeable noise and artifacts. Different from image denoising and super-resolution, in the field of MEF, the lack of ground truth is still the main stumbling block for deep learning. In addition, CNN-based methods require expensive GPU resources to store model parameters.
\begin{figure}
	\centering
	\includegraphics[width=1\linewidth]{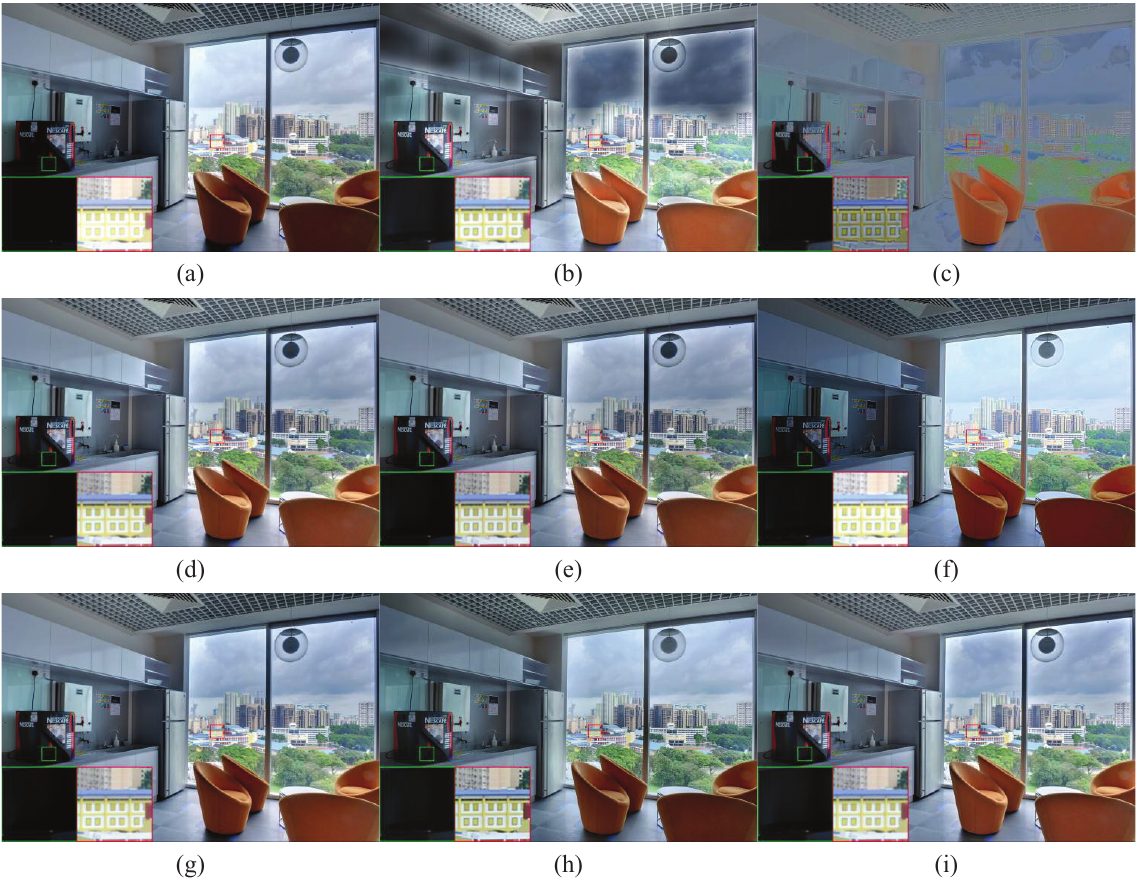}
	\vspace{-4mm}
	\caption{Example results of several different MEF methods. (a) Mertens09 \cite{Mertens_2009}. (b) S.Li13 \cite{Li_2013image}. (c) Shen14 \cite{Shen_2014}. (d) Kou17 \cite{kou2017multi}. (e) Z.Li17 \cite{Li_2017}. (f) Ma17 \cite{ma2017robust}. (g) Wang19 \cite{wang2019detail}. (h) Zhang19 \cite{zhang2020ifcnn}. (i) Proposed. Green and red boxes are zoom-in regions for taking a closer look. One can see that our method retains more details in the highlights and shadow regions compared to Mertens09\cite{Mertens_2009}.}
	\label{fig1}
	\vspace{-4mm}
\end{figure}

Although \cite{Mertens_2009} is simple and efficient, it suffers visible loss of detail in shadows and exposed areas (see Fig. \ref{fig1}(a)). The previous detail-enhancement components-based methods \cite{Kou_2018, Li_2012, Li_2017, kou2017multi, kou2017intelligent, wang2019detail} can add details to the final resulting image while the process is usually time-consuming and unfriendly to mobile devices, especially smart phones. In this article, we re-examine three classic exposure metrics \cite{Mertens_2009} and improve two of them. Under the LP framework, our method can maintain fine details in the highlights and capture better exposure in the shadow areas. Overall, our contributions are four-fold:
\begin{enumerate}
\item We propose a perceptual multi-exposure fusion method in the LP domain. Based on the difference between
varying exposure images and cross-channel correlation of color images, we reanalyze the potential flaws of three classic exposure metrics \cite{Mertens_2009} and improve two of them.
\item We find that traditional saturation definition is sensitive to grayscale content of color images. 
A adaptive well-exposedness (AWE) is designed in YCbCr color space, which considers the differences between images in an exposure sequence. The gradient of a color image (3-D gradient) is employed to extract fine details, which explores the cross-channel correlation of color imags.
\item This work constructs a large-scale multi-exposure dataset for static scenes that consists of 167 exposure sequences and covers a wide range of scenes. Experiments on the constructed dataset indicate that our method excels existing eight state-of-the-art approaches in term of both visual and quantitative results.
\item The proposed method can facilitate current image enhancement technologies, recording more detail in the highlight areas. All datasets and our MATLAB code are available at\footnote{\url{https://github.com/hangxiaotian/Perceptual-Multi-exposure-Image-Fusion}}.
\end{enumerate}
The rest of this paper is structured as follows. In section~\ref{sec:2}, we briefly review three classical exposure quality metrics and LP fusion framework. Section~\ref{sec:3} describes the proposed MEF approach in detail. Experimental results and evaluation are provided in Section~\ref{sec:4}. The final Section~\ref{sec:5} summarizes the paper and proposes further discussion. 
\section{Related Work}
\label{sec:2}
In this section, we review three classical exposure quality metrics \cite{Mertens_2009} and multi-scale fusion framework based on Laplace pyramid \cite{Burt_1983}. 
\subsection{Exposure Quality Metrics}
\subsubsection{Contrast}
The grayscale version of a color image is filtered with Laplace operator and then the absolute value of the filter response is taken as contrast denoted by $C$, which will give larger weight to the edges and textures than flat areas.
\subsubsection{Saturation}
Saturation can reflect over/under-exposed areas under most conditions, which is calculated by the standard deviation within the three channels, $R$, $G$, and $B$ at the corresponding pixel. Thus, saturation $S$ can be written as
\begin{equation}
\label{eq:eq1}
S{\rm{ = }}\sqrt {{{({I_R} - {I_\mu })}^2} + {{({I_G} - {I_\mu })}^2} + {{({I_B} - {I_\mu })}^2}},
\end{equation}
\begin{equation}
\label{eq:eq2}
{I_\mu }{\rm{ = }}\frac{1}{3}\left( {{I_R} + {I_G} + {I_B}} \right).
\end{equation}
where ${I_i}$ is the intensity on the $i$-\textit{th} channel, $i \in \left\{ {R,G,B} \right\}$, and ${I_\mu }$ denotes the corresponding mean of three channels.
\subsubsection{Well-exposedness}
Each pixel is weighted in the form of a Gaussian curve. The well-exposedness $E$ is defined as:
\begin{equation}
\label{eq:eq3}
E = {E_R} \cdot {E_G} \cdot {E_B},
\end{equation}
\begin{equation}
\label{eq:eq4}
{E_i} = \exp \left( { - \frac{{{{\left( {{I_i} - 0.5} \right)}^2}}}{{2{\sigma ^2}}}} \right){,i = R, G, B}.
\end{equation}
where ``$ \cdot $" denotes the pixel-wise multiplication and parameter $\sigma$ is set to 0.2.

According to above the three exposure quality measures \cite{Mertens_2009}, the initial weight map is given by,
\begin{equation}
\label{eq:eq5}
{W_n}\left( {x,y} \right) = {C_n}{\left( {x,y} \right)^{{w_C}}} \cdot {S_n}{\left( {x,y} \right)^{{w_S}}} \cdot {E_n}{\left( {x,y} \right)^{{w_E}}}.
\end{equation}
where $\left( {x,y} \right)$ and $n$ denote the location of pixel and $n$-\textit{th} input exposure image, respectively. ${w_C}$, ${w_S}$ and ${w_E}$ refer to the corresponding weighted indices, which default to 1.
\subsection{Multi-Scale Fusion Framework}
Firstly, to ensure space content consistency, each weight map is normalized by
\begin{equation}
\label{eq:eq6}
{\hat W_n}\left( {x,y} \right) = {\left[ {\sum\limits_{n = 1}^N {{W_n}\left( {x,y} \right)} } \right]^{ - 1}}{W_n}\left( {x,y} \right),
\end{equation}
The final resulting image $F$ is generated by
\begin{equation}
\label{eq:eq7}
F\left( {x,y} \right) = \sum\limits_{n = 1}^N {{{\hat W}_n}} \left( {x,y} \right){I_n}\left( {x,y} \right).
\end{equation}

Unfortunately, using Eq. (\ref{eq:eq7}) directly will cause seam problems. To address the issue, Mertens $\emph{et al}.$ \cite{Mertens_2009} adopted LP framework. LP is used to decompose input image sequence, and then pyramid images with different scales are blended. The steps are as follows: assumed that there are $N$ exposure images, and $N$ normalized weight maps served as alpha masks. Let $L{\left\{ A \right\}^l}$ represent that image $A$ is decomposed into $l$-\textit{th} layer ($l$$ \in \left\{ {1,2, \cdots ,L} \right\}$) in the LP decomposition, and analogously $G{\left\{ B \right\}^l}$ denotes the corresponding decomposed image of Gaussian pyramid. Thus, the pyramid is yielded by
\begin{equation}
\label{eq:eq8}
L{\left\{ {F\left( {x,y} \right)} \right\}^l} = \sum\limits_{n = 1}^N {G{{\left\{ {{{\hat W}_n}\left( {x,y} \right)} \right\}}^l}} L{\left\{ {{I_n}\left( {x,y} \right)} \right\}^l}.
\end{equation}
And finally, the Laplacian pyramid $L{\left\{ {F\left( {x,y} \right)} \right\}^l}$ is collapsed to generate the resulting image $F$.
\begin{figure}
	\centering
	\includegraphics[width=1\linewidth]{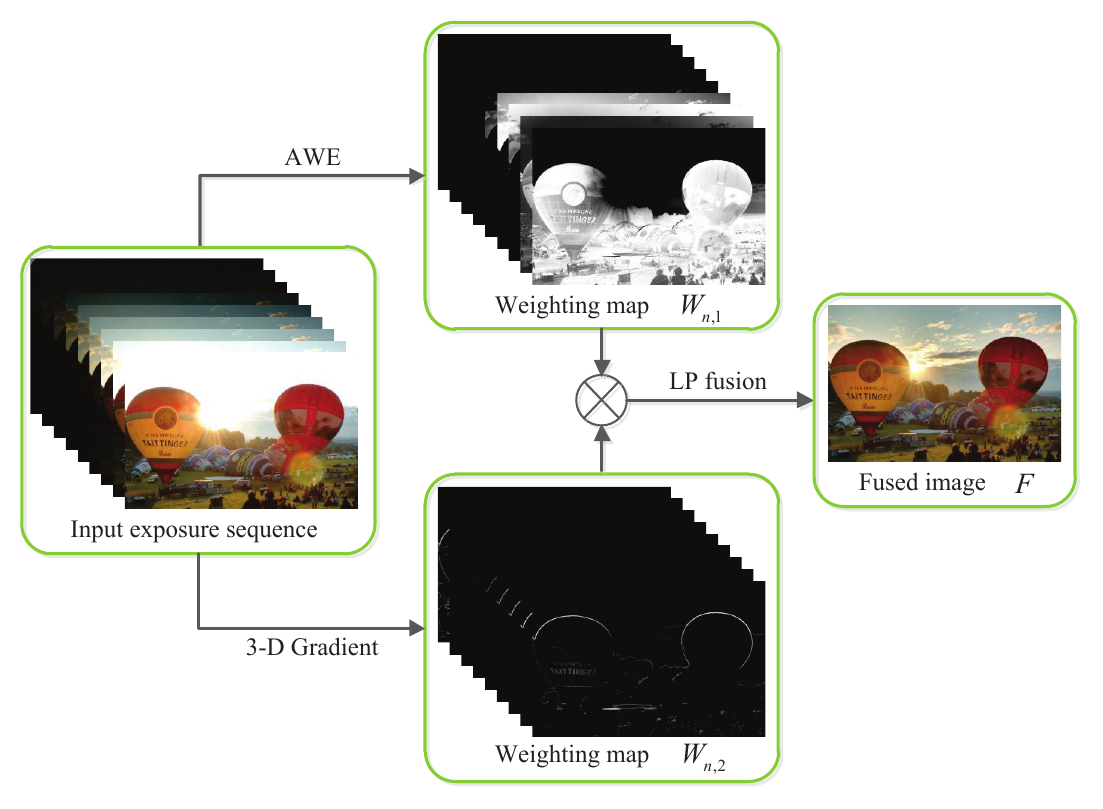}
	\vspace{-4mm}
	\caption{The pipeline of the proposed MEF method.}
	\label{fig2}
	\vspace{-4mm}
\end{figure}
\section{Proposed Multi-exposure Image Fusion}
\label{sec:3}
This section presents the proposed MEF method. We first analyze the defects of saturation, present an adaptive well-exposedness (AWE) and introduce the gradient of color images, i.e., 3-D gradient that is the generalization of conventional 2-D gradient. Then, we employ LP framework \cite{Burt_1983} to generate the final image. The pipeline of the proposed method is shown in Fig. \ref{fig2}.
\begin{figure}[htbp]
	\centering
	\includegraphics[width=0.3\linewidth]{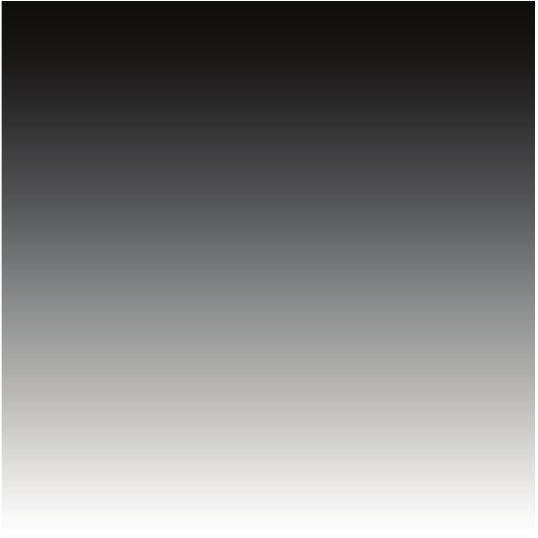}
	\vspace{-4mm}
	\caption{A synthetic color image with a size of $256 \times 256$ where the pixel value of three channels in each row is the same and increases line by line from 0 to 255.}
	\label{fig3}
	\vspace{-4mm}
\end{figure}
\begin{figure}[t]
	\centering
	\includegraphics[width=1\linewidth]{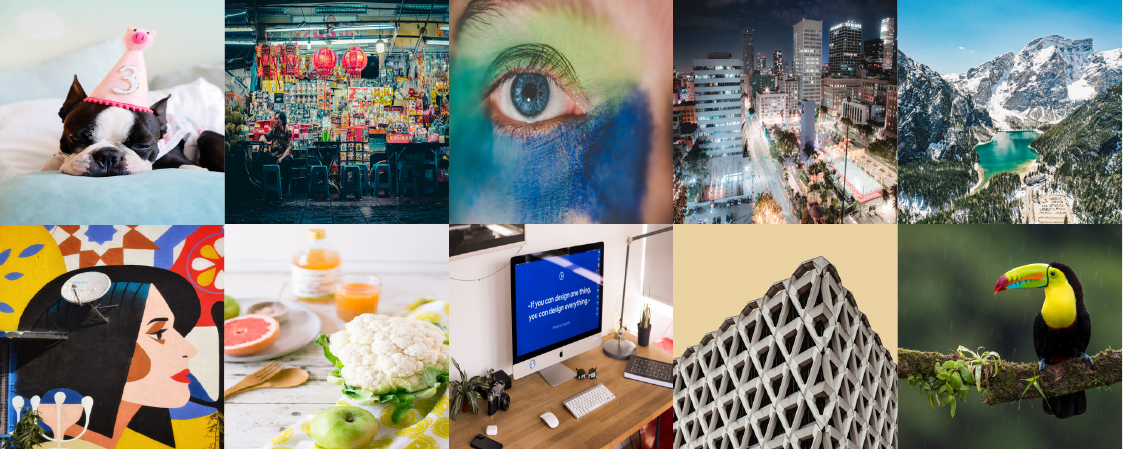}
	\vspace{-4mm}
	\caption{Some example images from our HRP dataset.}
	\label{fig4}
	\vspace{-4mm}
\end{figure}
\subsection{Rethinking Saturation}
We discover that although saturation can reflect the strength of exposure under certain conditions in which the grayscale content of color images is slight, the grayscale content of color images will be wrongly treated as over/under-exposed regions. This finding can be obtained from foregoing  Eqs. (\ref{eq:eq1}) and (\ref{eq:eq2}), i.e., when $R\left( {x,y} \right) = G\left( {x,y} \right) = B\left( {x,y} \right)$, saturation $S\left( {x,y} \right)$ is 0. In this case, saturation will mistake the grayscale content of color images as an over-exposed or under-exposed area.
For ease of understanding, Fig. \ref{fig3} shows a synthetic color image with a size of $256 \times 256$ where the pixel value of three channels in each row is the same and increases line by line from 0 to 255. According to Eqs. (\ref{eq:eq1}) and Fig. \ref{fig3}, we can thus obtain that saturation defined in \cite{Mertens_2009} is extremely vulnerable to the grayscale content of a color image. To overcome the defect and improve computational efficiency, saturation is not used in this work to design exposure metrics. 
\subsection{Adaptive Well-Exposedness}
We note that 0.5 in Eq. (\ref{eq:eq4}) is a constant, which does not take into account the differences between the exposed images, and can’t give much weight to those areas where pixel values are away from 0.5, such as in dark or bright areas. Therefore, it is desired to redesign this weighting function so as to be capable of recording more detail in bright/dark light. To this end, we naturally consider an image feature that could reveal luminance variation of different exposure images in the same sequence instead of 0.5. We naturally ask: ``which image feature can well uncover exposure changes"? Since image mean reflects the global brightness of exposure images and the mean with moderate exposure is approximate to 0.5, it leads us to utilize one as image characteristic. Let the mean of $n$-\textit{th} image is ${\mu _n}$, when ${I_n}\left( {x,y} \right)$ is close to $\left( {1 - {\mu _n}} \right)$, i.e., $\left( {{I_{\rm{n}}}\left( {x,y} \right) - \left( {1 - {\mu _n}} \right)} \right)$ is small, the corresponding pixel position should be assigned a larger value. Conversely, smaller weights should be given. Say, when an input image is overexposed, its mean ${\mu _n}$
is larger and thus $\left( {1 - {\mu _n}} \right)$ is smaller. At this point, the pixel value of the area with better exposure is small, so pixels close to $\left( {1 - {\mu _n}} \right)$ should be given greater weight.

Furthermore, we note that parameter $\sigma $ is set to 0.2 in Eq. (\ref{eq:eq4}). The reason for this setting is not explained in \cite{Mertens_2009}. Since the purpose of MEF is to meet the need of HDR photography, this reminds us whether parameter $\sigma $ can be set through statistical analysis of high-resolution photography (HRP) images. Thereupon, we construct a large-scale dataset, HRP, which contains 20,000 high-resolution images (most image resolution ranges from $4000 \times 3000$ to $8000 \times 6000$). 
The dataset is built via collecting freely useable images on website Unsplash\footnote{Available:  \url{https://unsplash.com/}}, which covers a variety of characteristic elements like textures, nature, architecture, animals, food, current events, travel, fashion, and so on. Since the resolution is too large, we display them via down-sampling. Some example images from our dataset are shown in Fig. \ref{fig4}. 
\begin{figure}
	\centering
	\includegraphics[width=1\linewidth]{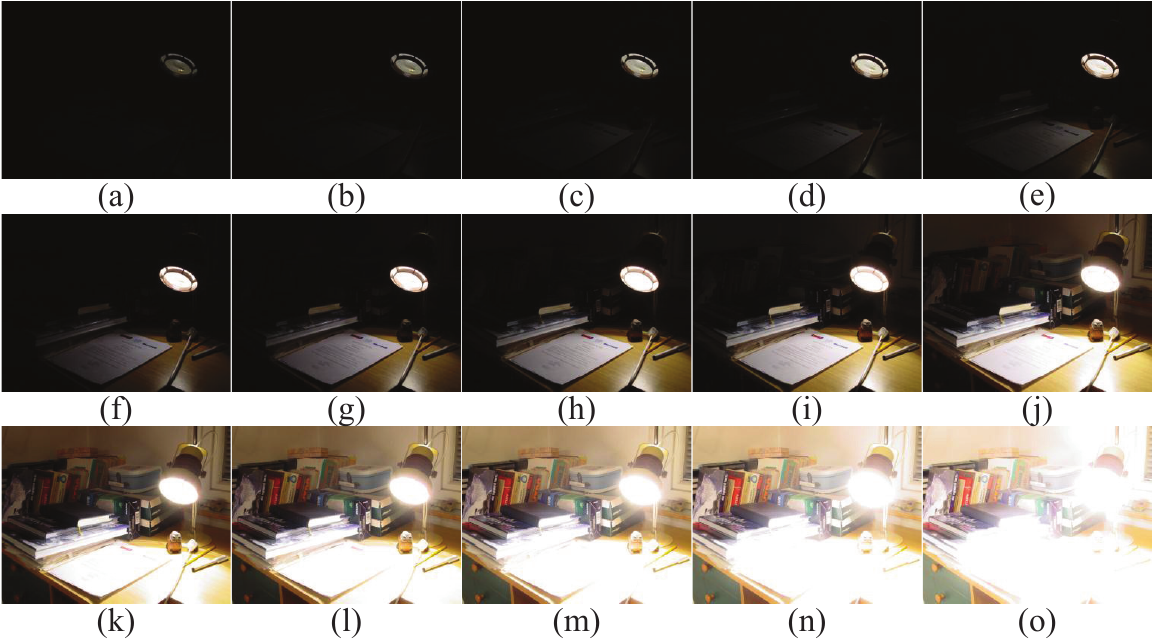}
	\vspace{-4mm}
	\caption{“Lamp” exposure sequence with different exposure time.}
	\label{fig5}
	\vspace{-4mm}
\end{figure}
\begin{figure}
	\centering
	\includegraphics[width=1\linewidth]{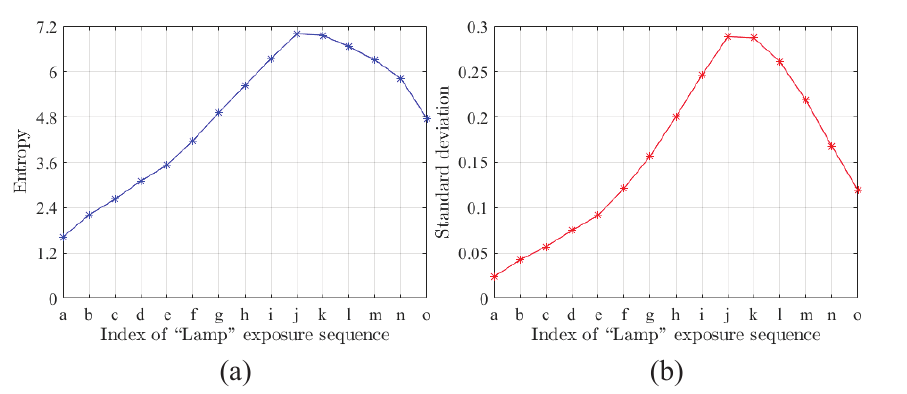}
	\vspace{-4mm}
	\caption{Statistical characteristics of ``Lamp" exposure sequence obtained on the 15 images of Fig. \ref{fig5}. (a) Entropy of exposure sequence. (b) Standard deviation of exposure sequence.}
	\label{fig6}
	\vspace{-4mm}
\end{figure}

Next, similarly, we will ask which image features can be well used to analyze the parameter $\sigma $? Inspired by \cite{kim2018exposure} and the density of a univariate Gaussian distribution, we consider using entropy and standard deviation to analyze parameter $\sigma $ in Eq. (\ref{eq:eq4}). Note that Eq. (\ref{eq:eq4}) is formally similar to the density function of the Gaussian distribution. Though entropy can reveal the amount of information, entropy in some conditions when the image content contains a large number of pure color areas, such as the sky of pure blue and the grass of pure green, is relatively small. Moreover, entropy may be more vulnerable to the pure color area of the image content than standard deviation. To support this viewpoint, we randomly take ``Lamp” exposure sequence obtained from \cite{ma2017robust} as an example in Fig. \ref{fig5}, calculating images entropy and standard deviation of the exposure sequence, respectively. From Figs. \ref{fig5} and \ref{fig6}, we see that while image entropy and standard deviation can present the strength of exposure, standard deviation is more sensitive to over-exposure because its curve tends to fall more sharply. In addition, the standard deviation is between $\left[ {0,1} \right]$, which is more in line with the range of parameter $\sigma$. Hence, the standard deviation is more preferable for analyzing parameter $\sigma $ compared to image entropy.
\begin{figure}
	\centering
	\includegraphics[width=0.5\linewidth]{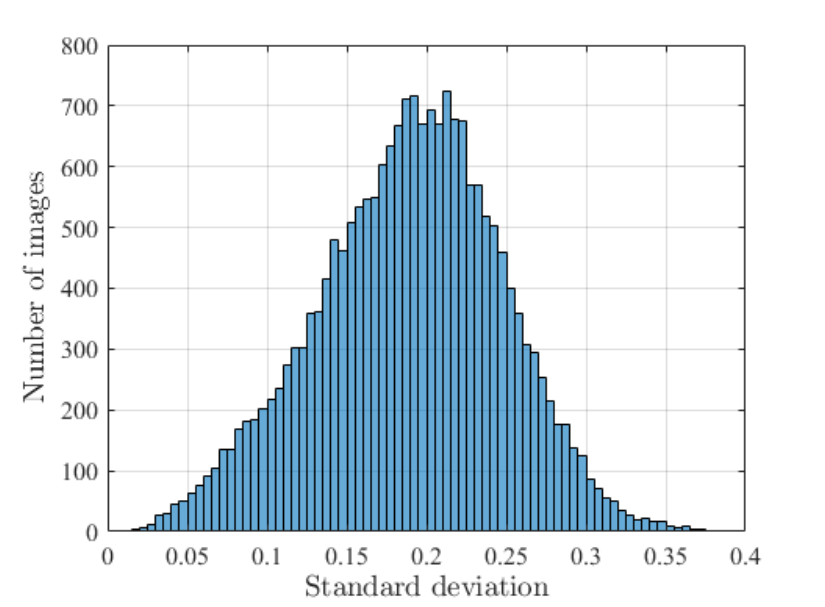}
	\vspace{-4mm}
	\caption{Histogram of standard deviation calculated from our HRP dataset.}
	\label{fig7}
	\vspace{-4mm}
\end{figure}
 Furthermore, we convert the image in the HPR dataset from RGB space to YCbCr space, and then use its luminance component to calculate the standard deviation. The statistical results of the standard deviation post in Fig. \ref{fig7}. The standard deviation is concentrated in $[0.15,0.25]$. To sum up, an adaptive well-exposure (AWE) is defined as follows:
\begin{equation}
\label{eq:eq9}
{W_{n,1}}\left( {x,y} \right) = \exp \left( { - \frac{{{{\left( {{I_{n,Y}}\left( {x,y} \right) - \left( {1 - {\mu _n}} \right)} \right)}^2}}}{{2{\sigma ^2}}}} \right).
\end{equation}
where ${I_{n,Y}}$ indicates Y channel of YCbCr color space of the $n$-\textit{th} image, i.e., luminance component, and ${\mu _n}$ refers to the mean of ${I_{n,Y}}$. In this work, parameter $\sigma$ is also set to 0.20 through Fig. \ref{fig7} and experimental analysis in detail (see the discussion of parameter $\sigma$ in Section~\ref{sec:4_B}).

\begin{figure}
	\centering
	\includegraphics[width=1\linewidth]{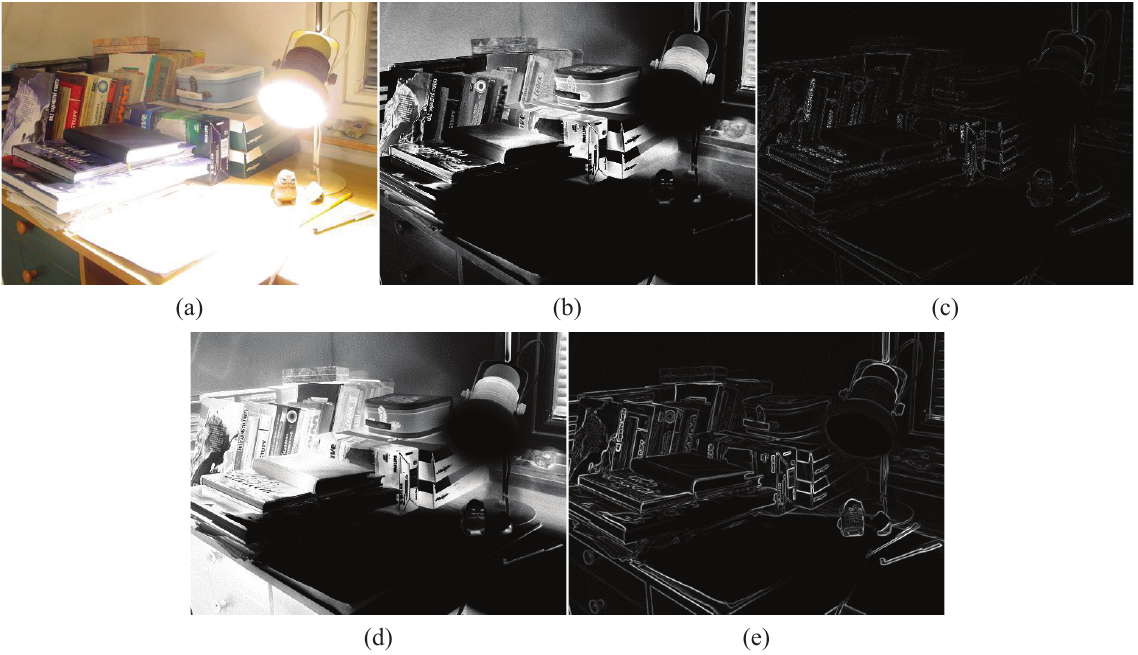}
	\vspace{-4mm}
	\caption{Comparison of our two weight maps with Mertens09's \cite{Mertens_2009}. (a) Over-exposed ``Lamp" image from the $m$-\textit{th} in Fig. \ref{fig5}. (b) and (c) are the weight maps defined in Mertens09 : Well-exposedness and Contrast, respectively. (d) and (e) are our weight maps: AWE and 3-D gradient magnitude, respectively.
	 }
	\label{fig8}
	\vspace{-4mm}
\end{figure}

Unlike the input of Eq. (\ref{eq:eq4}), to reduce the correlation among three channels and color distortion, we take the luminance component of YCbCr color space as input in lieu of an isolated input to RGB space \cite{Mertens_2009}. Moreover, the human visual perception is more sensitive to the Y component that well reflects the global illumination changes of exposure images. As shown in Fig. \ref{fig8} (a), (b) and (d), taking the $m$-\textit{th} over-exposed image in Fig. \ref{fig5} as an example, our AWE can capture more detailed information than Well-exposedness of Mertens09 \cite{Mertens_2009}.
\subsection{The Gradient of Color Images}
Even though contrast \cite{Mertens_2009} and gradient \cite{Zhang_2011} can effectively reveal over/under-exposed areas of a natural scene, most existing gradient based methods merely focus on traditional 2-D space. It's possible that the correlation among three channels is affected. There are three common ways to compute the gradient of color images. (1) The first strategy is directly to employ grayscale image as input. (2) The second strategy is to convert RGB space to another color space, such as YCbCr, YUV, and then compute the gradient of one component in transformed color space \cite{wang2019detail}. (3) The last strategy is to take the root mean square or sum of the gradients of the three channels. In all of these strategies the image-channels do not actually cooperate with one another, i.e., edge evidence along a given direction in one channel does not reinforce edge evidence along the same direction in other channels \cite{di1986note}. To avoid this, the use of the tensor gradient of multi-components regarded as vector field is adopted \cite{di1986note}.

A continuous color image can be regarded as a function mapping, i.e., $\boldsymbol{f}:{\Re ^2} \mapsto {\Re ^3}$, where $\Re$ denotes the set of real number. If $x$ and $y$ are two space coordinates, the color components can be denoted by $R\left( {x,y} \right)$, $G\left( {x,y} \right)$, and $B\left( {x,y} \right)$. Then, the color image will be written as $\boldsymbol{f} = \left( {R\left( {x,y} \right),G\left( {x,y} \right),B\left( {x,y} \right)} \right)$. The following notations will be adopted: $\left( {x,y} \right) = \left( {{x_1},{x_2}} \right) = \boldsymbol{x}$, $\boldsymbol{f} = \left( {R,G,B} \right) = \left( {{f_1},{f_2},{f_3}} \right)$, $\boldsymbol{y} = \boldsymbol{f}\left( \boldsymbol{x} \right) = \left( {{f_1}\left( x \right),{f_2}\left( x \right),{f_3}\left( x \right)} \right)$, $V = \left\{ {\boldsymbol{f}\left( \boldsymbol{x} \right),\boldsymbol{x} \in {\Re ^2}} \right\}$.

For $j = 1,2,3$, and $i = 1,2$, we assume that the rank of Jacobian matrix $\boldsymbol{J} = \left[ {{{\partial {f_j}} \mathord{\left/
			{\vphantom {{\partial {f_j}} {\partial {x_i}}}} \right.
			\kern-\nulldelimiterspace} {\partial {x_i}}}} \right]$
is 2 everywhere in ${\Re ^2}$. Then $V$ is a two-dimensional manifold embedded in ${\Re ^3}$. Let $\boldsymbol{{f_i}}\left( x \right) = \left( {{{\partial {f_1}} \mathord{\left/
			{\vphantom {{\partial {f_1}} {\partial {x_i}}}} \right.
			\kern-\nulldelimiterspace} {\partial {x_i}}},{{\partial {f_2}} \mathord{\left/
			{\vphantom {{\partial {f_2}} {\partial {x_i}}}} \right.
			\kern-\nulldelimiterspace} {\partial {x_i}}},{{\partial {f_3}} \mathord{\left/
			{\vphantom {{\partial {f_3}} {\partial {x_i}}}} \right.
			\kern-\nulldelimiterspace} {\partial {x_i}}}} \right)$, $i = 1,2$.
According to this definition, $\boldsymbol{{f_i}}\left( x \right)$ is a 3 tuple of real number. Moreover, we postulate that $\boldsymbol{{f_i}}\left( x \right)$ and its first derivatives are continuous. For $i,k = 1,2$, we set
\begin{equation}
\label{eq:eq10}
{g_{ik}}\left( \boldsymbol{x} \right) = \boldsymbol{{f_i}}\left( \boldsymbol{x} \right) \cdot \boldsymbol{{f_k}}\left( \boldsymbol{x} \right)
\end{equation}
where ``$ \cdot $" is dot product. Notice that ${\left\{ {\boldsymbol{{f_i}}\left( \boldsymbol{x} \right)} \right\}_{i = 1,2}}$ is a basis for the two-dimensional vector space of tangent vectors of $V$ at $\boldsymbol{y} = \boldsymbol{f}\left( \boldsymbol{x} \right)$. According to the above notations, the vectors $\boldsymbol{{f_i}}\left( x \right)$, $i = 1,2$ can be written
\begin{equation}
\label{eq:eq11}
\boldsymbol{u} = \frac{{\partial R}}{{\partial x}}\boldsymbol{r} + \frac{{\partial G}}{{\partial x}}\boldsymbol{g} + \frac{{\partial B}}{{\partial x}}\boldsymbol{b},
\end{equation}
\begin{equation}
\label{eq:eq12}
\boldsymbol{v} = \frac{{\partial R}}{{\partial y}}\boldsymbol{r} + \frac{{\partial G}}{{\partial y}}\boldsymbol{g} + \frac{{\partial B}}{{\partial y}}\boldsymbol{b}.
\end{equation}
where $\boldsymbol{r}$, $\boldsymbol{g}$, and $\boldsymbol{b}$ are unitary vectors associated with $R$, $G$, and $B$ axes, respectively. Similarly, ${\left\{ {{g_{ik}}\left( \boldsymbol{x} \right)} \right\}_{i,k = 1,2}}$ can be represented by
\begin{equation}
\label{eq:eq13}
{g_{xx}} = \boldsymbol{u} \cdot \boldsymbol{u} = {\left| {\frac{{\partial R}}{{\partial x}}} \right|^2} + {\left| {\frac{{\partial G}}{{\partial x}}} \right|^2} + {\left| {\frac{{\partial B}}{{\partial x}}} \right|^2},
\end{equation}
\begin{equation}
\label{eq:eq14}
{g_{yy}} = \boldsymbol{v} \cdot \boldsymbol{v} = {\left| {\frac{{\partial R}}{{\partial y}}} \right|^2} + {\left| {\frac{{\partial G}}{{\partial y}}} \right|^2} + {\left| {\frac{{\partial B}}{{\partial y}}} \right|^2},
\end{equation}
\begin{equation}
\label{eq:eq15}
{g_{xy}} = {g_{yx}} = \boldsymbol{u} \cdot \boldsymbol{v} = \frac{{\partial R}}{{\partial x}}\frac{{\partial R}}{{\partial y}} + \frac{{\partial G}}{{\partial x}}\frac{{\partial G}}{{\partial y}} + \frac{{\partial B}}{{\partial x}}\frac{{\partial B}}{{\partial y}}.
\end{equation}

In the application of image processing, we are often interested in the following two quantities \cite{di1986note} computed locally at each space coordinate $\left( {x,y} \right)$: (1) the direction through $\left( {x,y} \right)$ along which $\boldsymbol{f}$ has the maximum rate of change. (2) the absolute value of this maximum rate of change. Therefore, we aim to find the maximization of the following form
\begin{equation}
\label{eq:eq16}
{\left( {d\boldsymbol{f}} \right)^2} = {g_{xx}}dxdx + {g_{yy}}dydy + {g_{xy}}dxdy + {g_{yx}}dydx,
\end{equation}
subject to
\begin{equation}
\label{eq:eq17}
dxdx + dydy = 1.
\end{equation}
The above problem can also be formulated as the following optimization form
\begin{equation}
\label{eq:eq18}
\mathop {\arg \max }\limits_\theta  {\rm{ }}{g_{xx}}{\cos ^2}\theta  + 2{g_{xy}}\cos \theta \sin \theta  + {g_{yy}}{\sin ^2}\theta.
\end{equation}
Let $F\left( \theta  \right) = {g_{xx}}{\cos ^2}\theta  + 2{g_{xy}}\cos \theta \sin \theta  + {g_{yy}}{\sin ^2}\theta $. Using the common trig function formulas: ${\sin ^2}\theta  = {{\left( {1 - \cos 2\theta } \right)} \mathord{\left/
		{\vphantom {{\left( {1 - \cos 2\theta } \right)} 2}} \right.
		\kern-\nulldelimiterspace} 2}$, ${\cos ^2}\theta  = {{\left( {1 + \cos 2\theta } \right)} \mathord{\left/
		{\vphantom {{\left( {1 + \cos 2\theta } \right)} 2}} \right.
		\kern-\nulldelimiterspace} 2}$ and $\sin \theta \cos \theta  = {{\sin 2\theta } \mathord{\left/
		{\vphantom {{\sin 2\theta } 2}} \right.
		\kern-\nulldelimiterspace} 2}$, $F\left( \theta  \right)$ can thus be updated to
\begin{equation}
\label{eq:eq19}
\begin{aligned}
F\left( \theta  \right) &= \frac{1}{2}\left[ {{g_{xx}}\left( {1 + \cos 2\theta } \right) + 2{g_{xy}}\sin 2\theta  + {g_{yy}}\left( {1 - \cos 2\theta } \right)} \right] \\ 
&= \frac{1}{2}\left[ {\left( {{g_{xx}} + {g_{yy}}} \right) + \left( {{g_{xx}} - {g_{yy}}} \right)\cos 2\theta  + 2{g_{xy}}\sin 2\theta } \right]. \\ 
\end{aligned}
\end{equation}
Let ${{dF} \mathord{\left/
		{\vphantom {{dF} {d\theta }}} \right.
		\kern-\nulldelimiterspace} {d\theta }} = 0$, one can obtain
\begin{equation}
\label{eq:eq20}
\theta \left( {x,y} \right) = \frac{1}{2}\arctan \left( {\frac{{2{g_{xy}}}}{{{g_{xx}} - {g_{yy}}}}} \right).
\end{equation}
Indeed, $\theta \left( {x,y} \right)$ is the angle which determines the direction through point $\left( {x,y} \right)$ along which $f$ has the maximum rate of change. Consider $\tan \left( \alpha  \right) = \tan \left( {\alpha  \pm \pi } \right)$, if ${\theta _0}$ is the solution to the above equation, so is ${\theta _0} \pm {\pi  \mathord{\left/
		{\vphantom {\pi  2}} \right.
		\kern-\nulldelimiterspace} 2}$. 
As $F\left( \theta  \right) = F\left( {\theta  + \pi } \right)$
, we merely need to compute the values of $\theta $ within the interval $\left[ {0,\pi } \right)$. Eq. (\ref{eq:eq19}) has two values ${\pi  \mathord{\left/
		{\vphantom {\pi  2}} \right.
		\kern-\nulldelimiterspace} 2}$ 
apart at each point $\left( {x,y} \right)$, which means that involves a pair of orthogonal directions: along one of them, $\boldsymbol{f}$ attains its maximum rate of change, along the other, its minimum. Therefore the absolute value of this maximum rate of change is given by
\begin{equation}
\label{eq:eq21}
\begin{aligned}
\begin{array}{c}
{G_\theta }\left( {x,y} \right) = \left\{ {\frac{1}{2}\left[ {\left( {{g_{xx}} + {g_{yy}}} \right) + \left( {{g_{xx}} - {g_{yy}}} \right)\cos 2\theta \left( {x,y} \right) + } \right.} \right. \\ 
{\left. {\left. {2{g_{xy}}\sin 2\theta \left( {x,y} \right)} \right]} \right\}^{{1 \mathord{\left/
				{\vphantom {1 2}} \right.
				\kern-\nulldelimiterspace} 2}}} \\ 
\end{array}
\end{aligned}
\end{equation}
where ${G_\theta }\left( {x,y} \right)$ denotes the gradient magnitude at coordinate $\left( {x,y} \right)$.

The above derivation is based on the fact that $\boldsymbol{f}$ is a continuous color image. For a digital color image, however, \cite{di1986note} used a surface fitting technique to approximate the gradient. Please refer to \cite{di1986note} for more derivation of numerical approximations of $\theta \left( {x,y} \right)$ and ${G_\theta }\left( {x,y} \right)$.

We can use the \textbf{Sobel} operator to calculate the partial derivative in the above equation. Finally, the gradient magnitude of a color image (3-D gradient) serves as the second weight:
\begin{equation}
\label{eq:eq22}
{W_{n,2}}\left( {x,y} \right) = {G_{n,\theta }}\left( {x,y} \right).
\end{equation}
where ${G_{n,\theta }}\left( {x,y} \right)$ denotes the gradient magnitude of the $n$-\textit{th} color image at coordinate $\left( {x,y} \right)$.

\begin{figure}
	\centering
	\includegraphics[width=1\linewidth]{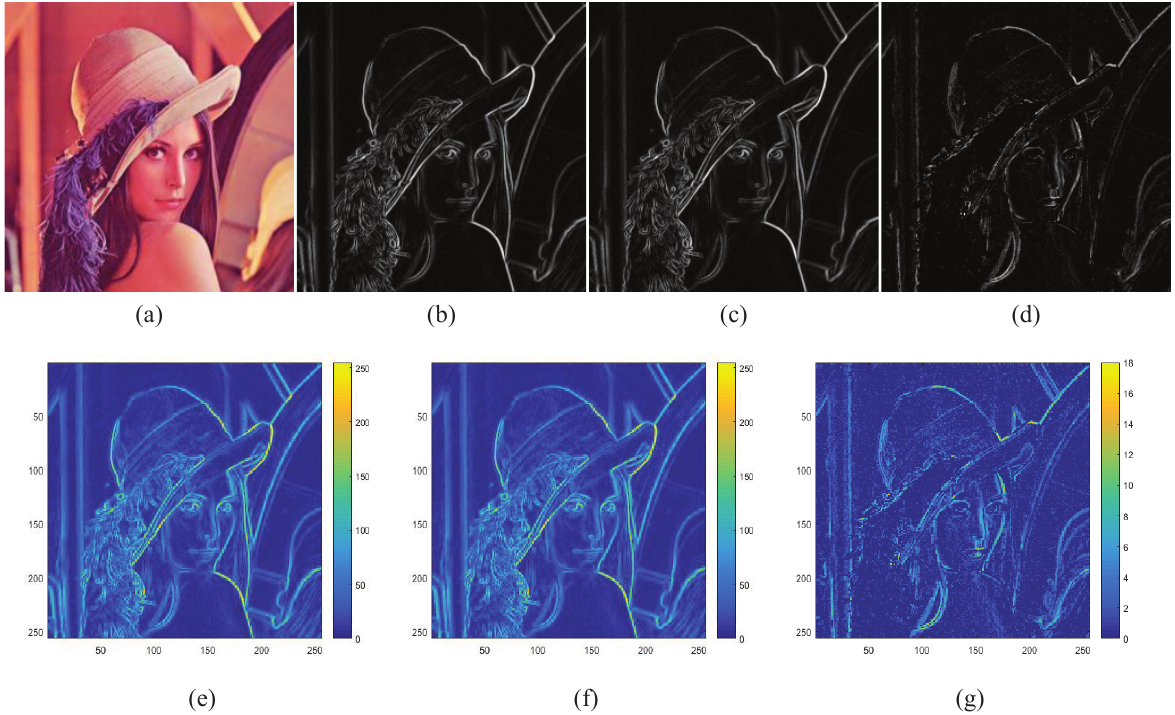}
	\vspace{-4mm}
	\caption{Comparison between 3-D gradient and traditional 2-D gradient. (a) Input image ``Lena" . (b) 3-D gradient computed in RGB space. (c) The sum of the gradient computed on each channel. (d) The difference between (b) and (c). (e)-(g) Visualization from (b) to (d).
	}
	\label{fig9}
	\vspace{-4mm}
\end{figure}

As shown in Fig. \ref{fig9}, the edge details of the vector gradient (3-D gradient) are more complete than the ones that are the sum of the gradient images of the three channels. Say, the difference in visualization can be easily observed from Fig. \ref{fig9} (e)-(g). Moreover, Fig. \ref{fig8} (c) and (e) shows a comparison of contrast defined in \cite{Mertens_2009} with 3-D gradient. Both indicate that the 3-D gradient can extract more fine details than the contrast or the traditional 2-D gradient.
\subsection{Multi-Scale Fusion Framework}
After combining the two weights above, i.e., ${W_{n,1}}\left( {x,y} \right)$ and ${W_{n,2}}\left( {x,y} \right)$, the initial weight map can be given by,
\begin{equation}
\label{eq:eq23}
{W_n}\left( {x,y} \right) = {W_{n,1}}{\left( {x,y} \right)^{{w_1}}} \cdot {W_{n,2}}{\left( {x,y} \right)^{{w_2}}}.
\end{equation}
where ``$ \cdot $" is the pixel-wise multiplication and ${w_1}$, ${w_2}$ are positive numbers, which emphasize the importance of the corresponding weight map.

Since we scale the input image pixels into $\left[ {0,1} \right]$, the weight maps calculated by Eqs. (\ref{eq:eq9}) and (\ref{eq:eq22}) are also within the range. Due to the limited dynamic range of imaging sensors, noise artifacts frequently occur in extremely low/dark light. In such cases, noise is often mistakenly considered by as a change in image contrast. On the other hand, there are smaller gradients in smooth or less texture areas, and vice versa. To mitigate the impact of the gradient on these areas, ${w_2}$ should be assigned larger weights. We therefore set ${w_1} = 1,{w_2} = 2.2$ (see the discussion of parameter ${w_2}$ in Section~\ref{sec:4_B}).

Furthermore, to reduce the impact of outliers and noise artifacts, the final weight map ${W_{n,g}}\left( {x,y} \right)$ is smoothed by using Gaussian filter:
\begin{equation}
\label{eq:eq24}
{W_{n,g}}\left( {x,y} \right) = GF\left( {{W_n}\left( {x,y} \right),{\sigma _g}} \right),
\end{equation}
where $GF\left(  \cdot  \right)$ denotes Gaussian filter operator, and ${\sigma _g}$ is standard deviation of Gaussian smoothing kernel, and set to 3 (see the discussion of parameter ${\sigma _g}$ in Section~\ref{sec:4_B}).

In order to maintain spatial consistency, we normalize each weight map by
\begin{equation}
\label{eq:eq25}
{\hat W_n}\left( {x,y} \right) = {\left( {\sum\limits_{n = 1}^N {{W_{n,g}}\left( {x,y} \right)}  + \varepsilon } \right)^{ - 1}}{W_{n,g}}\left( {x,y} \right).
\end{equation}
where $\varepsilon $ is a very small constant in case the denominator is 0.

At the end, we employ LP based framework \cite{Mertens_2009} (using Eq. (\ref{eq:eq8})) to generate the final fusion image.
\section{Experiments}
\label{sec:4}
In this section, we first present experimental settings and parameters analysis of the proposed method. Then, we provide an experimental comparison with existing eight state-of-the-art technologies on the built MEF dataset. We further investigate our method with a notable improvement in detail preservation compared to current image enhancement technologies. Finally, the running time of different methods is compared.
\begin{table}[t]
	\centering
	\setlength{\belowcaptionskip}{-0.cm}
	\caption{The constructed MEF benchmark dataset for static scenes.}
	\begin{tabular}{m{4em}<{\centering} |c |c |c |c |c}
		\hline
		Dataset & dataset 1 & dataset 2 & dataset 3 & dataset 4 & dataset 5 \\
		\hline
		\rule{0pt}{3pt}
		\thead{Number of\\ sequence}  & 44    & 46    & 33    & 24    & 20 \\
		\hline
	\end{tabular}%
	\label{tab1}%
\end{table}%
\subsection{Experimental Settings}
\subsubsection{MEF Datasets}
\label{sec:4_A_1}
To evaluate the performance of current state-of-the-art methods, we implement a large number of experiments on five datasets for the first time, which contains 167 multi-exposure sequences for static natural scenes in total. Table \ref{tab1} lists the number of exposure sequences for each dataset, please refer to the link below for more information. Exposure images of dataset 1\footnote{Available:  \url{http://rit-mcsl.org/fairchild/HDR.html}} and dataset 3\footnote{Available: \url{http://www.empamedia.ethz.ch/hdrdatabase/index.php}} captured in camera RAW format and then we convert them to TIF format by utilizing Photoshop2018 while some information may be lost in the process. There are two sequences in dataset 3 that are not the camera RAW format but the JPEG format. Since the size of exposure images in the dataset 1 and dataset 3 are too large as well as the low efficiency of some methods such as \cite{Shen_2014, ma2017robust, zhang2020ifcnn}, we down-sample ones directly by a factor of 8.  Dataset 2\footnote{Available:  \url{http://val.serc.iisc.ernet.in/DeepFuseICCV17/}} is enthusiastically provided by the authors of \cite{ram2017deepfuse}. In \cite{yang2018adaptive}, dataset 3 was used to investigate HDR adjustment. Dataset 4\footnote{Available:  \url{https://ece.uwaterloo.ca/~k29ma/}} has been widely used in many MEF approaches \cite{Li_2017, kou2017multi, kou2017intelligent, ma2015perceptual, wang2019detail, ma2015multi, ma2017robust, ram2017deepfuse, li2018multi}. Dataset 5 is collected from previous literatures and Internet.
\subsubsection{High-Resolution Photography Dataset}
\label{sec:4_A_2}
To analyze parameter $\sigma $ in Eq. (\ref{eq:eq9}), we construct a large-scale high-resolution photography (HRP) dataset that consists of 20,000 high-resolution wallpapers (most images with the size of $4000 \times 3000$ to $8000 \times 6000$) freely accessed from Unsplash, which is the internet’s source of freely usable images powered by creators everywhere, and covers a variety of scenarios, such as textures, nature, animals, films, architectures, current events, travels, food, fashion, spirituality and the like. Initially, we collect more than 40,000 images. Many images are broken or duplicated, so it relies on a heavy manual process to choose carefully. After the selection process, there are only over 20,000 images left, and then we randomly select 20,000 photos to be the HRP dataset. Some example images from our HRP dataset are shown in Fig. \ref{fig4}. Since the purpose of MEF is to meet the need of photography, the HRP dataset we built is different from the previous dataset, such as \cite{ma2016group}. The HRP dataset constructed from the perspective of photography is more in line with the related tasks of mobile photography, such as aesthetics evaluation of consumer photography, image denoising, super-resolution, and MEF. In particular, each image of the HRP dataset can produce a multi-exposure sequence using gamma correction and contrast or brightness adjustment. After cropping, we can use them to train an end-to-end learned system.  This leads us to research in the next work.
\begin{table}[t]
	\setlength{\abovecaptionskip}{0.cm}
	\setlength{\belowcaptionskip}{-0.cm}
	\centering
	\caption{Parameters of the proposed method}
	\begin{tabular}{c|c|c|c|c|c}
		\hline
		Parameters & $\sigma $     & ${w_1}$     & ${w_2}$     & ${\sigma _g}$     &$L$  \\
		\hline
		Corresponding value & 0.2   & 1  & 2.2   & 3   & Eq. (\ref{eq:eq21}) \\
		\hline
	\end{tabular}%
	\label{tab2}%
\end{table}
\begin{figure}
	\centering
	\includegraphics[width=1\linewidth]{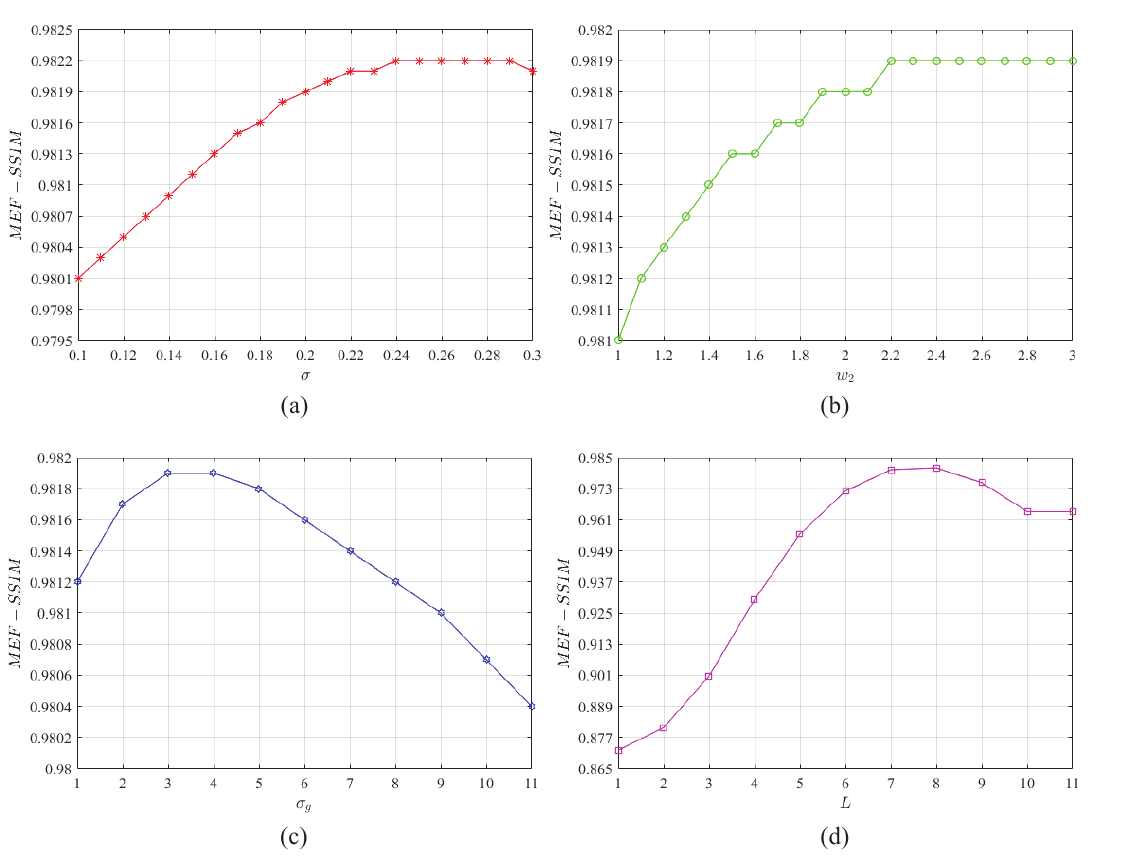}
	\vspace{-4mm}
	\caption{Impact of our method’s parameters on objective performance.}
	\label{fig10}
	\vspace{-4mm}
\end{figure}
\begin{figure*}[!htbp]
	\centering
	\includegraphics[width=1\linewidth]{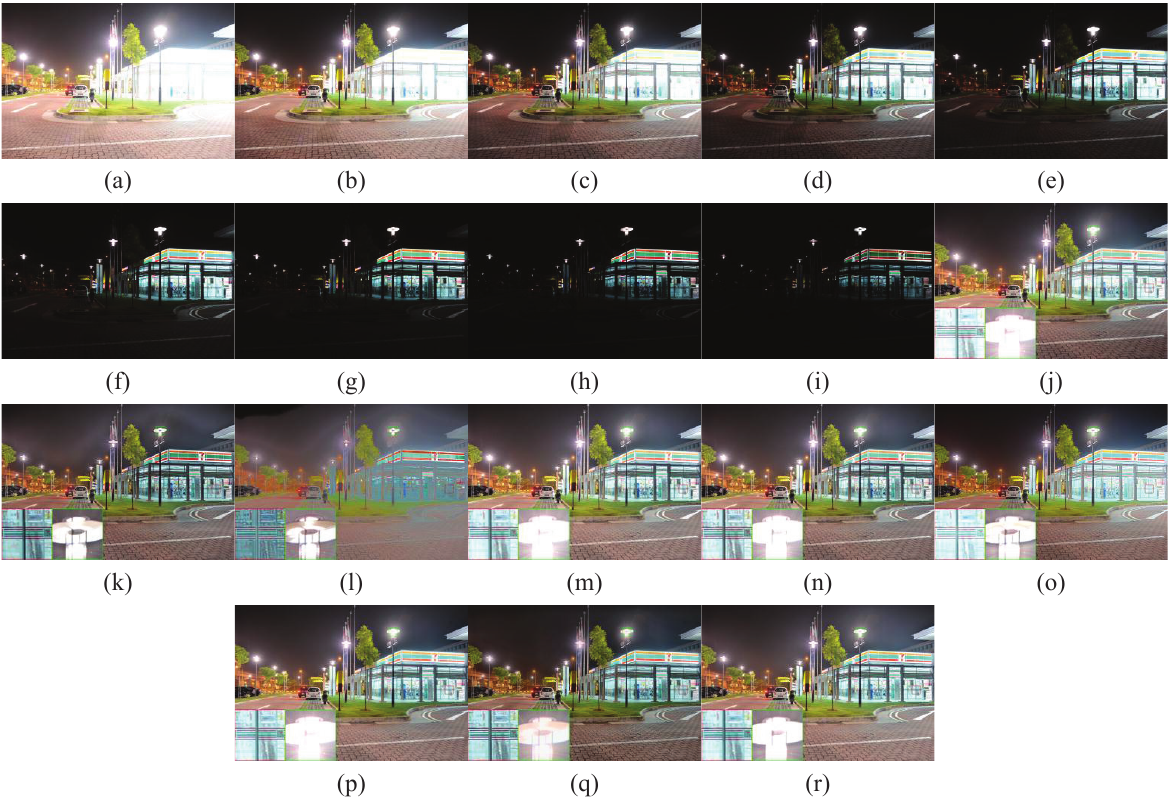}
	\vspace{-4mm}
	\caption{Visual results of different MEF methods on the image sequence “Seven Night”. (a)-(i) Input multi-exposure sequences. (j) Mertens09 \cite{Mertens_2009}. (k) S.Li13 \cite{Li_2013image}. (l) Shen14 \cite{Shen_2014}. (m) Kou17 \cite{kou2017multi}. (n) Z.Li17 \cite{Li_2017}. (o) Ma17 \cite{ma2017robust}. (p) Wang19 \cite{wang2019detail}. (q) Zhang19 \cite{zhang2020ifcnn}. (r) Proposed. Red and Green boxes are zoom-in regions for taking a closer look.}
	\label{fig11}
	\vspace{-4mm}
\end{figure*}
\begin{figure*}[htbp]
	\centering
	\includegraphics[width=1\linewidth]{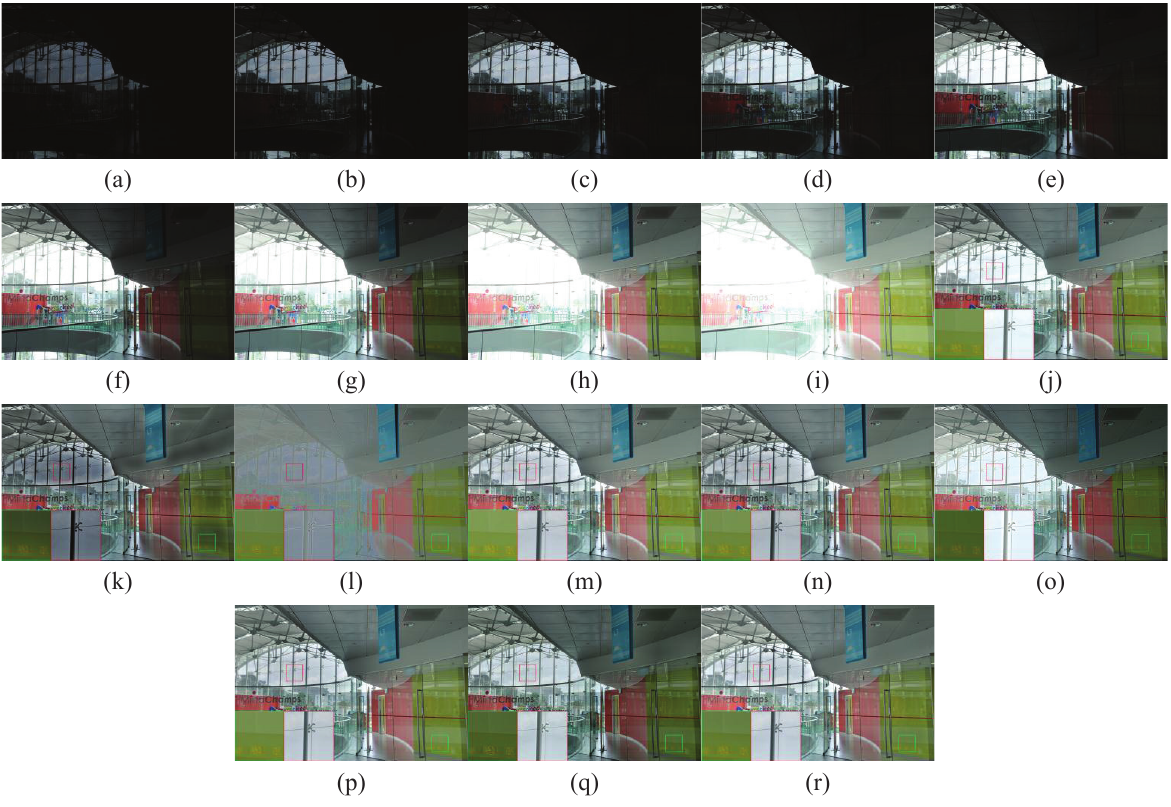}
	\vspace{-4mm}
	\caption{Visual results of different MEF methods on the image sequence ``Preschool". (a)-(i) Input multi-exposure sequences. (j) Mertens09 \cite{Mertens_2009}. (k) S.Li13 \cite{Li_2013image}. (l) Shen14 \cite{Shen_2014}. (m) Kou17 \cite{kou2017multi}. (n) Z.Li17 \cite{Li_2017}. (o) Ma17 \cite{ma2017robust}. (p) Wang19 \cite{wang2019detail}. (q) Zhang19 \cite{zhang2020ifcnn}. (r) Proposed. Red and Green boxes are zoom-in regions for taking a closer look.}
	\label{fig12}
	\vspace{-4mm}
\end{figure*}
\subsubsection{Comparison with State-of-the-art MEF Methods}
\label{sec:4_A_3}
We compare the proposed perceptual MEF (PMEF) method against existing eight competitive approaches: Mertens09 \cite{Mertens_2009} (CGF’09), S.Li13 \cite{Li_2013image} (TIP’13), Shen14 \cite{Shen_2014} (TCYB’14), Kou17 \cite{kou2017multi} (ICME’17), Z.Li17 \cite{Li_2017} (TIP’17), Ma17 \cite{ma2017robust} (TIP’17), Wang19 \cite{wang2019detail} (TCSVT’19), Zhang19 \cite{zhang2020ifcnn} (INFFUS’19), on the built MEF benchmark dataset. The letters and numbers in parentheses indicate the journal abbreviation and the year published officially, respectively. The codes of the Mertens09\footnote{Available:  \url{https://mericam.github.io/}}, S.Li13\footnote{Available:  \url{http://xudongkang.weebly.com/index.html}}, Shen14\footnote{Available:  \url{https://github.com/shenjianbing/BLPfusion}}, Kou17\footnote{Available:  \url{http://koufei.weebly.com/}}, Z.Li17\footnote{Available:  \url{https://github.com/weizhe/deef}}, Ma17\footnote{Available:  \url{https://ece.uwaterloo.ca/~k29ma/}}, Wang19\footnote{Available:  \url{https://github.com/QTWANGBUAA/exposure-fusion}}, and Zhang19\footnote{Available:  \url{https://github.com/uzeful/IFCNN}} based methods are available online. The parameters for all compared methods are set by default values. It is worth noting in \cite{Shen_2014} that when $N\left( {N \ge 4} \right)$ exposure images are inputted, ${\gamma _0} = {\left[ {1, \cdots ,1} \right]_{1 \times N}}$, otherwise ${\gamma _0} = \left[ {0.8,1,1.2} \right]$.
\subsubsection{Objective Assessment Metrics}
\label{sec:4_A_4}
MEF-SSIM \cite{ma2015perceptual} derived from structural similarity and patch structural consistency is an image quality assessment (IQA) model against static scene. The quality of finally fused images is proportional to its score.

\subsubsection{Experimental Environment}
\label{sec:4_A_5}
Aside from the experiment of Zhang19 \cite{zhang2020ifcnn} (see in Section~\ref{sec:5_F} for details), other experiments are conducted on MATLAB R2016b installed on Win 10 64-bit operating system and a PC with Intel Core i7-6700K CPU (4.0 GHz), 32 GB RAM.
\subsection{Parameters Analysis}
\label{sec:4_B}
There are five free parameters in our method in which are reported Table \ref{tab2}. Note that unlike the setting of layer  for LP decomposition in \cite{Burt_1983, kou2017multi, wang2019detail}, in this paper the layer $N$ is set to
\begin{equation}
\label{eq:eq26}
L = \begin{cases}
7, & \text{if }  \textit{$N$} > 3,\\
8, & \text{others}.
\end{cases}
\end{equation}
where $N$ indicates the number of input exposure images. We have found through experiments that when multiple images are taken in the same exposure sequence, the exposure difference between adjacent images is small. At this time, if the number of decomposition layers is large, the final result is easily over-exposed. Except for parameter ${w_1}$ defaults to 1, other parameters are analyzed by using controlling for a variable, only varying one parameter while others are fixed. In particular, parameters analysis experiments are performed 24 sets of images in the dataset 4, which is the most widely used one in MEF community. Here, we adopt MEF-SSIM \cite{ma2015perceptual} to analyze the influence of these parameters.

Fig. \ref{fig10} indicates the change of MEF-SSIM while varying each of parameters. From Fig. \ref{fig10}(a), as parameter $\sigma $ increases, the score of MEF-SSIM first gradually rises and then nearly remains stable till $\sigma {\rm{ = }}0.29$. In spite of the overall trend similar to parameter $\sigma $, the growth of parameter ${w_2}$ is slightly significant in Fig. \ref{fig10}(b). We see that parameter $\sigma $ selected in this paper is not the best ones for all exposure sequences on dataset 4 while it can obtain competitive results with smaller fluctuations on the other datasets. Through performing numerous experiments against five datasets above, we found that the average score of the MEF-SSIM is very close when parameter $\sigma $ belongs to the range $[0.2,0.24]$. In addition, to highlight the advantages of the proposed method over \cite{Mertens_2009}, we therefore set parameter $\sigma {\rm{ = }}0.2$ to be the same as that of \cite{Mertens_2009}. For Fig. \ref{fig10}(c), the score of MEF-SSIM shows a significant growth between ${\sigma _g}{\rm{ = }}1$ and ${\sigma _g}{\rm{ = }}3$, rising to almost 0.9819. The score, however, stays momently stable and then declines steadily to 0.9804. This reveals that the selected Gaussian smoothing kernel is effective for all image sequences. As can be seen from Fig. \ref{fig10}(d), the setting strategy of parameter $L$, i.e., Eq.~(\ref{eq:eq21}), is reasonable as the tradeoff between efficiency and performance. In addition, we make numerous experiments suggesting that the selection of these parameters is not dramatically critical to the quantitative results of five datasets above. The error is very small as long as the setting of parameters is close to what is set in Table \ref{tab2}.
\begin{table*}[h]
	\centering
	\caption{Quantitative results with MEF-SSIM (higher is better) on five datasets. Each row is highlighted with red, green and blue to indicate the top first three results, respectively. }
	\begin{tabular}{c|ccccccccc}
		\hline
		Dataset & Mertens09 \cite{Mertens_2009} & S.Li13 \cite{Li_2013image} & Shen14 \cite{Shen_2014} & Kou17 \cite{kou2017multi} & Z.Li17 \cite{Li_2017} & Ma17 \cite{ma2017robust}  & Wang19 \cite{wang2019detail} & Zhang19 \cite{
			zhang2020ifcnn} & Proposed \\
		\hline
		Dataset 1 & \textcolor[rgb]{ 0,  .69,  .314}{\textbf{0.9814}} & 0.9707  & 0.6951  & 0.9723  & 0.9698  & \textcolor[rgb]{ 0,  .439,  .753}{\textbf{0.9769}} & 0.9746  & 0.9462  & \textcolor[rgb]{ 1,  0,  0}{\textbf{0.9830}} \\
		Dataset 2 & \textcolor[rgb]{ 1,  0,  0}{\textbf{0.9856}} & 0.9708  & 0.8808  & 0.9823  & 0.9817  & 0.9807  & \textcolor[rgb]{ 0,  .439,  .753}{\textbf{0.9826}} & 0.9486  & \textcolor[rgb]{ 0,  .69,  .314}{\textbf{0.9849}} \\
		Dataset 3 & \textcolor[rgb]{ 0,  .69,  .314}{\textbf{0.9679}} & 0.9470  & 0.6822  & 0.9524  & 0.9513  & \textcolor[rgb]{ 0,  .439,  .753}{\textbf{0.9651}}  & 0.9574 & 0.9340  & \textcolor[rgb]{ 1,  0,  0}{\textbf{0.9698}} \\
		Dataset 4 & \textcolor[rgb]{ 0,  .69,  .314}{\textbf{0.9790}} & 0.9677  & 0.7655  & 0.9709  & 0.9689  & 0.9743  & \textcolor[rgb]{ 0,  .439,  .753}{\textbf{0.9753}} & 0.9270  & \textcolor[rgb]{ 1,  0,  0}{\textbf{0.9819}} \\
		Dataset 5 & \textcolor[rgb]{ 0,  .69,  .314}{\textbf{0.9720}} & 0.9550  & 0.7505  & \textcolor[rgb]{ 0,  .439,  .753}{\textbf{0.9636}} & 0.9621  & 0.9633  & 0.9625  & 0.9244  & \textcolor[rgb]{ 1,  0,  0}{\textbf{0.9748}} \\
		\hline
		Average & 0.9772  & 0.9622  & 0.7548  & 0.9683  & 0.9668  & 0.9721  & 0.9705  & 0.9360  & 0.9789  \\
		\hline
		Rank  & 2     & 7     & 9     & 5     & 6     & 3     & 4     & 8     & 1  \\
		\hline
	\end{tabular}%
	\label{tab3}%
\end{table*}%
\subsection{Visual Comparison}
Fig. \ref{fig11} shows visual comparison results of different methods on the image sequence ``Seven Night". Although Shen14 \cite{Shen_2014} retains rich color information in over-exposed areas, the overall contrast is somewhat reduced. The image as a whole shows a misty feeling (see Fig. \ref{fig11}(l)). Mertens09 \cite{Mertens_2009}, Wang19 \cite{wang2019detail}, Z.Li17 \cite{Li_2017}, and Kou17 \cite{kou2017multi} exhibit different degrees of over-exposedness at the street lamp (zoom in for improved visibility). Ma17 \cite{ma2017robust} and Mertens09 \cite{Mertens_2009} suffers from some halo artifacts in glass areas (see green zoom-in regions). Compared with the above method, the proposed method can retain fine detail information and color information in the over-exposed area. While Zhang19 \cite{zhang2020ifcnn} introduces some color noise at the street lamp, it captures rich color information in the glass area. 
In general, S.Li \cite{Li_2013image} can well preserve the details in the brightest and darkest regions and the color saturation well as shown in Fig. \ref{fig11}(k) .
Although the color information generated by Shen11 [47] is slightly better, but the overall
contrast is somewhat reduced. Shen14 \cite{Shen_2014} ensures good exposure in the highlights such as sky, but also shows serious and evident noise artifacts that are not quite as impressive in indoor conditions (see Fig. \ref{fig11}(i)). 

Fig. \ref{fig12} shows visual comparison results of different methods on the image sequence ``Preschool". As is shown in Fig. \ref{fig12}(k), even though there are salient blackish color casts visible like indoor windows, glasses as well as ceiling, S.Li \cite{Li_2013image} can record well-preserved detail in the highlight regions. Fig. \ref{fig12}(k) evidently underexposes the subject, while Fig. \ref{fig12}(q) is capable of achieving fine exposure in indoor conditions. However, the only drawback to Zhang19 \cite{zhang2020ifcnn} is a slight lack of contrast especially on the ceiling. In contrast, as for Fig. \ref{fig12}(o), the level of captured image detail is pretty low when shooting in bright and dark light. Some other artifacts, moreover, including ringing and chromatic aberrations, can be visible in Fig. \ref{fig12}(o). From Fig. \ref{fig12}(l), Shen14 \cite{Shen_2014} still does a poor job of retaining detail, sharpness, exposure and color, even performing worse than other competitors in pretty much all light conditions. As for detail, Fig. \ref{fig12}(j) is not as plentiful as images captured with some methods such as Fig. \ref{fig12}(m), (n), (p), and (r). Overall, there is good detail in both the dark foreground area and the bright background for Fig. \ref{fig12}(m), (n), (p), and (r).
\begin{figure}[htbp]
	\centering
	\includegraphics[width=1\linewidth]{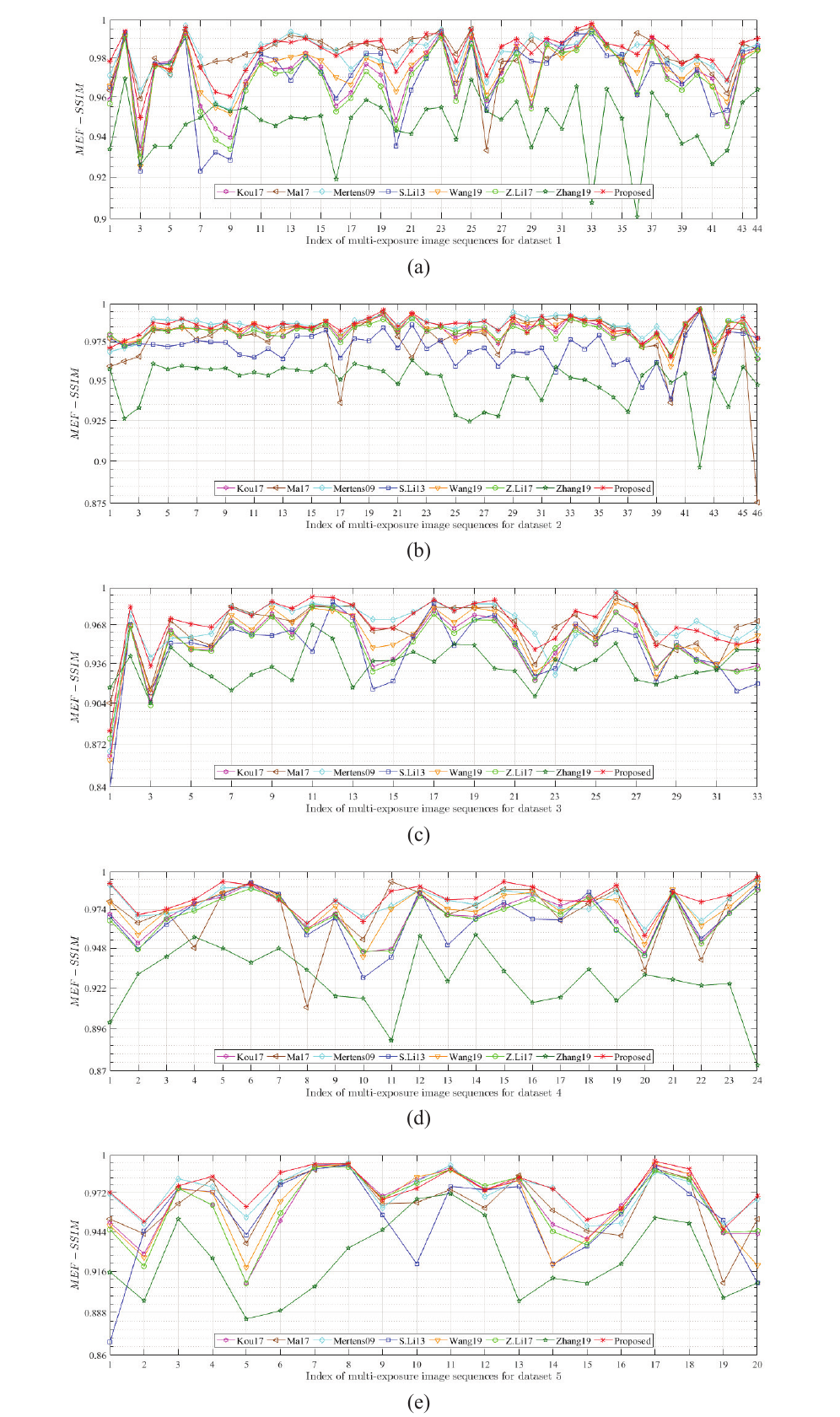}
	\vspace{-4mm}
	\caption{Quantitative results of different methods for five datasets. Please enlarge to see the difference.}
	\label{fig13}
	\vspace{-4mm}
\end{figure}
\subsection{Quantitative Evaluation}
Quantitative results with MEF-SSIM on five datasets are listed in Table \ref{tab3}. The average score of each method on each dataset is reported. Each row is highlighted with \textcolor{red}{\textbf{red}}, \textcolor[rgb]{ 0,  .69,  .314}{\textbf{green}} and \textcolor{blue}{\textbf{blue}} to indicate the top first three results, respectively. To rank all the methods, we give a comprehensive average score based on the average score of each method on each dataset. Our method achieves the highest scores among 4 datasets, and ranks second in dataset 2, just 0.0007 behind Mertens09 \cite{Mertens_2009}, which convincingly verify the effectiveness of proposed method. The score of Mertens09 \cite{Mertens_2009} follows our method closely. For the remaining methods, the top three are Ma17 \cite{ma2017robust}, Wang19 \cite{wang2019detail}, and Kou17 \cite{kou2017multi}, respectively. It is evident from Table \ref{tab3} that Shen14 \cite{Shen_2014} ranks the worst in overall datasets, which is consistent with the visual results in Figs. \ref{fig11}(l) and \ref{fig12}(l).
\begin{figure}[!t]
	\vspace{-2mm}
	\centering
	\begin{minipage}[b]{1\linewidth}
		{\includegraphics[width= 1\textwidth]{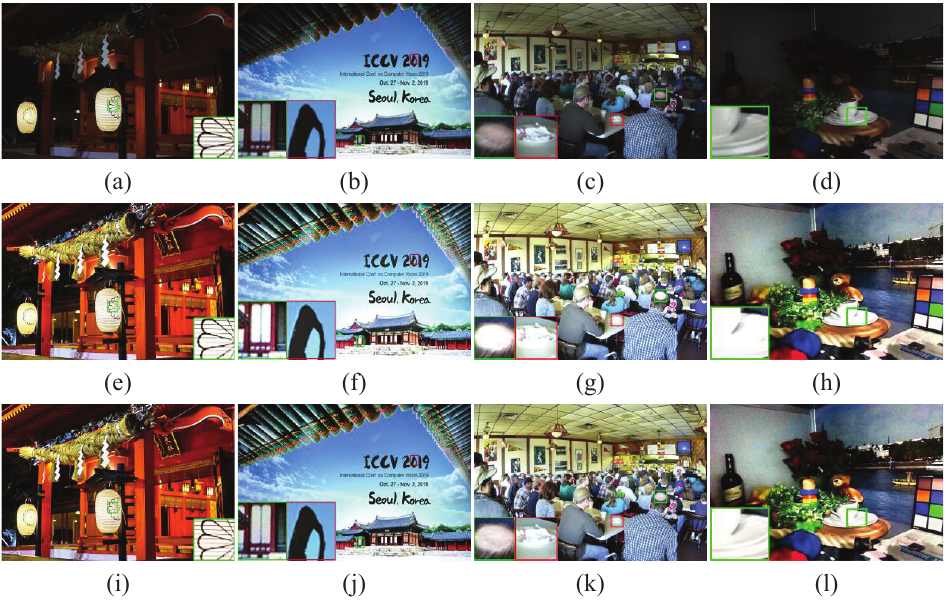}}
	\end{minipage}
	\vspace{-4mm}
	\caption{The application of the proposed method in image enhancement. (a)-(d) Input images: nighttime image, natural image, low-contrast image and low-light image, respectively. (e)-(h) The enhanced results by LIME. (i)-(l) The enhanced results by LIME+Proposed, which takes images corresponding to the first and second rows as input for multi-exposure fusion.}
	\label{fig14}
	\vspace{-4mm}
\end{figure}
As shown in Fig. \ref{fig13}, we visualize the quantitative results of different methods on each dataset to objectively evaluate them. From Table \ref{tab3}, we see that the results of Shen14 \cite{Shen_2014} are the worst on all datasets. For the best visualization, we only make a show of the quantitative results of other methods except for Shen14 \cite{Shen_2014}. It is evident from Fig. \ref{fig13}(a), (b), (d) and (e), the MEF-SSIM of Zhang19 \cite{zhang2020ifcnn} fluctuates wildly as exposure sequence varies. This is mainly due to the unpleasant noise artifacts in the resulting image of Zhang19 \cite{zhang2020ifcnn}. In addition to Zhang19 \cite{zhang2020ifcnn}, the MEF-SSIM of both S.Li13 \cite{Li_2013image} and Ma17 \cite{ma2017robust} on five datasets are also not stable enough. It is obvious that the score of Ma17 \cite{ma2017robust} is mainly affected by individual outliers. From the overall trend of each approach across all datasets, the performances of both our method and Mertens09 \cite{Mertens_2009} on dataset 2 are neck and neck, and our method is the top performer on the remaining four datasets.
\begin{table}[t]
	\centering
	\caption{Quantitative results with NIQE (low is better) and BIQME (high is better) obtained on the 4 images of Fig. \ref{fig14}.}
	\begin{tabular}{|c|c|c|c|c|c|c|}
		\hline
		\multirow{2}*{Image} & \multicolumn{2}{c|}{Input} & \multicolumn{2}{c|}{LIME} & \multicolumn{2}{c|}{LIME+Proposed} \\ 
		\cline{2-7}          & NIQE&BIQME&NIQE&BIQME&NIQE&BIQME \\
		\hline
		a     & 2.6080  & 0.3830  & 3.3955  & 0.5747  & 3.3283  & 0.5746  \\
		\hline
		b     & 2.4904  & 0.6486  & 2.3487  & 0.6463  & 2.3086  & 0.6473  \\
		\hline
		c     & 2.3387  & 0.5734  & 2.2740  & 0.6579  & 2.1322  & 0.6611  \\
		\hline
		d     & 2.9469  & 0.3212  & 2.8087  & 0.6166  & 2.4089  & 0.6193  \\
		\hline
	\end{tabular}%
	\label{tab4}%
\end{table}%
\subsection{Image Enhancement}
Most of image enhancement approaches are prone to overexpose those areas where exposure is normal or appropriate. In this subsection, we apply the proposed MEF method to improve enhanced results which obtain by using state-of-the-art low-light image enhancement (LIME) technology \cite{guo2016lime}. As shown in Fig.~\ref{fig14}(a)-(d), there are four types of input images, i.e., nighttime image, natural image, low-contrast image and low-light image, respectively. We combine source images and enhanced results by LIME as input images, and then use our MEF method to generate final resulting image. This strategy, LIME+Proposed, can improve image quality in terms of both quantitative and qualitative results. 
\subsubsection{Qualitative Comparison}
Although LIME improves the overall contrast, the level of captured image detail is slightly low and the enhanced images visibly overexposure in the highlights (see Fig. \ref{fig14}(e)-(h)). However, our strategy, LIME+Proposed, is a capable of high levels of detail (see Fig. \ref{fig14}(i)-(l)). 
\subsubsection{Quantitative Comparison}
We use two lately developed no-reference image quality assessment (NRIQA) measures, natural image quality evaluator (NIQE) \cite{mittal2012making} and blind image quality measure of enhanced images (BIQME) \cite{gu2017learning}, to evaluate the performance of image enhancement. NIQE that aims to assess the naturalness of an image is the most widely adopted quantitative measure in current image enhancement works. BIQME is a newly proposed NRIQA model for image enhancement, which is trained by abundant samples. Table \ref{tab4} reports their objective values obtained on the 4 images of Fig. \ref{fig14}(a)-(d). It is obvious that our strategy, LIME+Proposed, improves on both quality indicators compared to LIME. This full demonstrates the effectiveness of MEF technology in image enhancement.
\begin{table*}
	\centering
	\caption{The average runtime of compared methods on "Cades Cave" exposure sequence with 9 images at different scales.}
	\begin{tabular}{|c|c|c|c|c|c|c|}
		\hline
		\multirow{2}*{Method} & \multicolumn{6}{c|}{Runtime (seconds)} \\
		\cline{2-7}          & $134 \times 89$ & $268 \times 178$ & $536 \times 356$ & $1072 \times 712$ & $2144 \times 1424$ & $4288 \times 2848$ \\
		\hline
		Mertens09 \cite{Mertens_2009} & 0.0622  & 0.1880  & 0.6082  & 2.2791  & 8.7479  & 35.3560  \\
		\hline
		S.Li13 \cite{Li_2013image} & $\backslash$   & 0.1625  & 1.2535  & 4.7293  & 18.4378  & 79.3203  \\
		\hline
		Shen14 \cite{Shen_2014} & 1.6031  & 5.9318  & 24.7674  & 95.3120  & 368.6096  & $ \verb|\\|$  \\
		\hline
		Kou17 \cite{kou2017multi} & 0.0988  & 0.2564  & 1.3457  & 5.6653  & 22.9816  & 93.5699  \\
		\hline
		Z.Li17 \cite{Li_2017} & 0.2906  & 3.5049  & 0.9453  & 15.5934  & 65.5265  & 299.9742  \\
		\hline
		Ma17 \cite{ma2017robust}  & 0.4002  & 1.5781  & 5.9129  & 24.5470  & 99.8308  & 547.7291  \\
		\hline
		Wang19 \cite{wang2019detail}& 0.1054  & 0.2579  & 0.8934  & 3.4987  & 13.5214  & 53.4339  \\
		\hline
		Zhang19 \cite{zhang2020ifcnn} & 0.4667  & 1.6511  & 6.3766  & 25.2919  & 106.4605  & $ \verb|\\|$  \\
		\hline
		Proposed & 0.0908  & 0.2304  & 0.9909  & 3.8108  & 14.9255  & 59.6533  \\
		\bottomrule
	\end{tabular}%
	\label{tab5}%
\end{table*}%
\subsection{Runtime Comparison}
\label{sec:5_F}
We also compare the runtime of various methods and all experiments are performed in the environment aforesaid (see in Section~\ref{sec:4_A_5} for details). A “Cades Cave" exposure sequence is randomly selected from dataset 1, including 9 varying exposure images with the size of $4288 \times 2848$. For complete comparison, the exposure sequence is down-sampled directly by a factor 2, 4, 8, 16, 32, respectively. Since Zhang19 \cite{zhang2020ifcnn} is executed in PyTorch framework and a Titan-X GPU, we only utilize the CPU measure for fair comparison. Table~\ref{tab5} compares the mean execution time of different methods on ``Cades Cave" sequence with 9 images at different scales, which is calculated by carrying out 10 times tests. Due to the filter size of ${r_1} = 45$ in S.Li13 \cite{Li_2013image}, the index exceeds matrix dimension when input sequence size of $134 \times 89$. We use ``$\backslash$" to indicate corresponding time incalculable. Moreover, when input exposure size of $4288 \times 2848$, the implementation of Shen14 \cite{Shen_2014} and Zhang19 \cite{zhang2020ifcnn} indicates insufficient memory space. Here, we use ``$ \verb|\\|$" to denote out of memory. These two approaches therefore are the most time-consuming compared to others. Mertens09 \cite{Mertens_2009} becomes the top-ranked method for mean runtime in the sequence. The proposed method ranks third in our test, just slightly behind Wang19 \cite{wang2019detail}. The reason is that the calculation of 3-D gradient takes more time. 
\section{Conclusion and Future Work}
\label{sec:5}
An effective multi-scale exposure image method has been presented. Unlike current multi-scale fusion approaches, which have elaborately developed a detail-enhancement component, our pay attention to improve two new exposure quality metrics: adaptive well-exposedness (AWE) and the gradient of color images (3-D gradient). Through previous literatures and our collection, we create a novel large-scale multi-exposure dataset against static scenes. Experimental results demonstrate that the proposed method favorably outperforms eight competitive approaches like CNN \cite{zhang2020ifcnn} in both visually and MEF-SSIM scores. Our approach is well mobile-friendly compared to those methods that add a detail-enhanced component. We also create a high-resolution photographic (HRP) dataset, which is used to analyze exposure parameter of AWE and expected to promote research on related topics such as aesthetic quality assessment and HDR imaging. In addition, our approach can contribute to current image enhancement techniques, preserving good detail in the highlights in high-contrast conditions.  

There are some urgent problems addressed in the MEF community. (1) Image alignment must be considered, since the inevitable shaking occurs when users take pictures with a hand-held camera without tripod, especially for smartphones. (2) Removing ghosting from dynamic scenes is the central task of current research, especially at night. (3) Even though a new quality assessment for dynamic scenes, MEF-SSIMd \cite{fang2019perceptual}, has recently been proposed, it is still worth exploring the unification of quality assessment for static and dynamic scenes \cite{tursun2015state, tursun2016objective}.


%



%
%

\ifCLASSOPTIONcaptionsoff
  \newpage
\fi



%


\bibliographystyle{IEEEtran}
\bibliography{PMEF}

%








\end{document}